\documentclass[10pt,twocolumn,letterpaper]{article}

\usepackage{iccv}
\usepackage{times}
\usepackage{epsfig}
\usepackage{graphicx}
\usepackage{amsmath}
\usepackage{amssymb}

\usepackage{multirow}
\usepackage{makecell}
\usepackage{subfigure}
\usepackage{amsfonts}
\usepackage{bm}
\usepackage{booktabs}
\usepackage{bbding}
\usepackage{amsmath}
\usepackage{pdfpages}

\DeclareRobustCommand{\name}{TOOD~}

\DeclareRobustCommand{\head}{T-head~}

\usepackage[breaklinks=true,bookmarks=false]{hyperref}

\iccvfinalcopy 

\def\iccvPaperID{****} 

\ificcvfinal\fi

\begin{document}

\title{TOOD: Task-aligned One-stage Object Detection}

\author{Chengjian Feng$^{*}$\\
Intellifusion Inc. \\
{\tt\small feng.chengjian@intellif.com}
\and
Yujie Zhong$^{*}$\\
Meituan Inc.\\
{\tt\small zhongyujie@meituan.com}
\and
Yu Gao\\
ByteDance Inc.\\
{\tt\small gaoyu.1520@bytedance.com}
\and
Matthew R. Scott\\
Malong LLC \\
{\tt\small mscott@malongtech.com}
\and
Weilin Huang$^\dagger$\\
Alibaba Group \\
{\tt\small weilin.hwl@alibaba-inc.com}
}

\maketitle
\ificcvfinal\thispagestyle{empty}\fi

\begin{abstract}
     One-stage object detection is commonly implemented by optimizing two sub-tasks: object classification and localization, using heads with two parallel branches, 
     which might lead to a certain level of spatial misalignment in predictions between the two tasks.
     In this work, we propose a Task-aligned One-stage Object Detection (TOOD) that explicitly aligns the two tasks in a learning-based manner.
    	First, we design a novel Task-aligned Head (T-Head) which offers a better balance between learning task-interactive and task-specific features, as well as a greater flexibility to learn the alignment via a task-aligned predictor.
    	Second, we propose Task Alignment Learning (TAL) to explicitly pull closer (or even unify) the optimal anchors for the two tasks during training via a designed sample assignment scheme and a task-aligned loss.
    	Extensive experiments are conducted on MS-COCO, where TOOD achieves a \textbf{51.1 AP} at single-model single-scale testing. This surpasses the recent one-stage detectors by a large margin, such as ATSS~\cite{zhang2020bridging} (47.7 AP), GFL~\cite{li2020generalized} (48.2 AP), and PAA~\cite{kim2020probabilistic} (49.0 AP), with fewer parameters and FLOPs.
    	Qualitative results also demonstrate the effectiveness of TOOD for better aligning the tasks of object classification and localization.
    	Code is available at~\url{https://github.com/fcjian/TOOD}.
		\vspace{-6mm}
\end{abstract}

\newcommand\blfootnote[1]{%
\begingroup 
\renewcommand\thefootnote{}\footnote{#1}%
\addtocounter{footnote}{-1}%
\endgroup 
}
{
	\blfootnote{
	 $^*$Equal contributions. 
	 $^\dagger$ Corresponding author.
	 }
}

\section{Introduction}  \label{Introduction}
Object detection aims to localize and recognize objects of interest from natural images, and is a fundamental yet challenging task in computer vision. It is commonly formulated as a multi-task learning problem by jointly optimizing object classification and localization~\cite{feng2021exploring,girshick2014rich,jiang2018acquisition,lin2017focal,redmon2016you,zhong2020representation}.
	The classification task is designed to learn discriminative features that focus on the key or salient part of an object, while the localization task works on precisely locating the whole object with its boundaries.
	Due to the divergence of learning mechanisms for classification and localization, spatial distributions of the learned features by the two tasks can be different, causing a certain level of misalignment when predictions are made by using two separate branches.
	
\begin{figure}[t]
		\centering
		\small
		\setlength{\tabcolsep}{2.1mm}{
        \renewcommand\arraystretch{0.97}
            \begin{tabular}{c}
                \includegraphics[height=5.2cm]{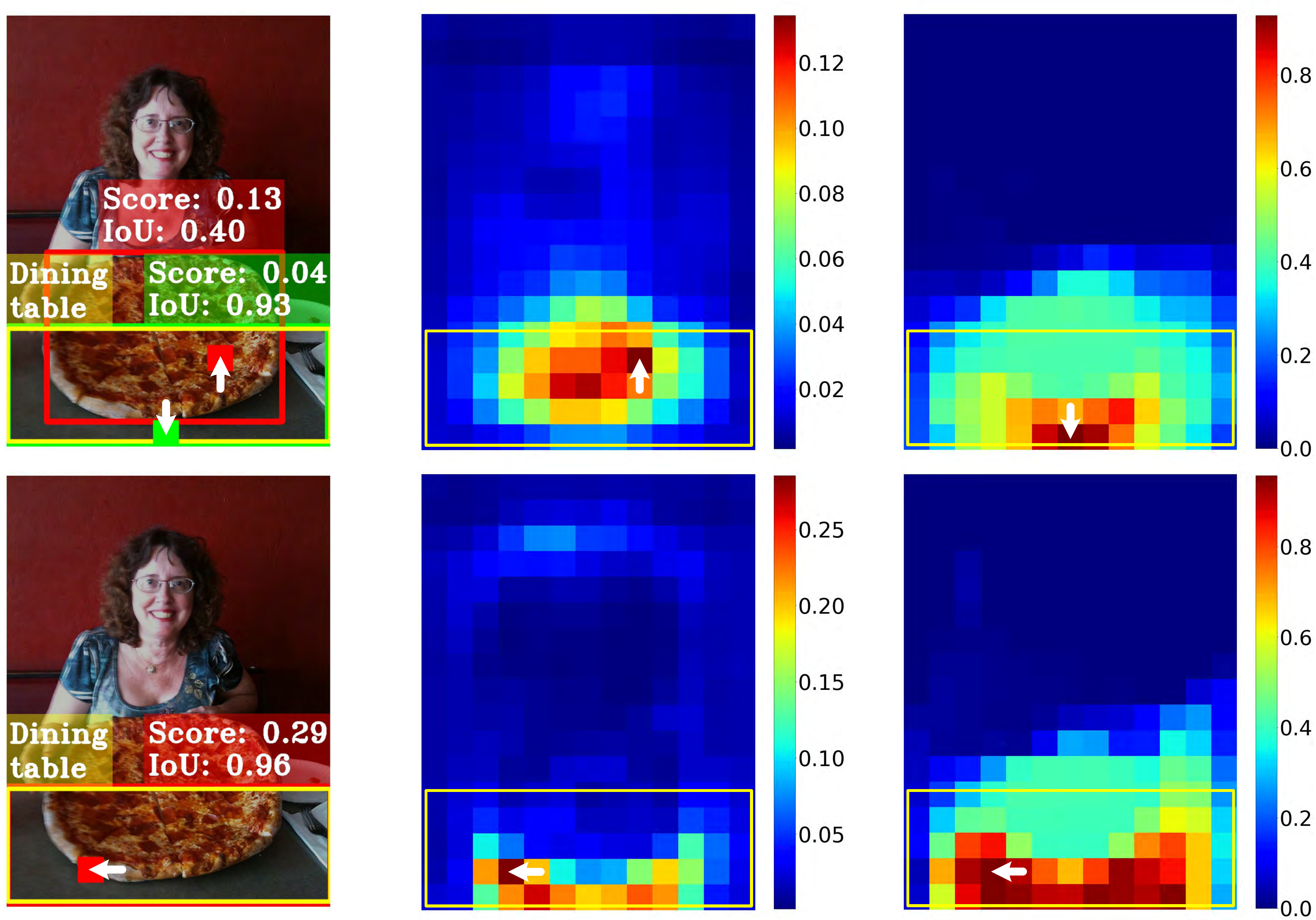}
                \\
                \quad Result \quad \quad \quad \quad \quad Score \quad \quad \quad \quad \quad \quad \quad IoU \quad \quad \quad
            \end{tabular}
		}
        \caption{Illustration of detection results~(`Result') and spatial distributions of classification scores~(`Score') and localization scores~(`IoU') predicted by ATSS~\cite{zhang2020bridging}~(top row) and the proposed TOOD~(bottom row). Ground-truth is indicated by yellow boxes, and a white arrow means the main direction of the best anchor away from the center of an object. In the `Result' column, a red/green patch is the location of the best anchor for classification/localization, while a red/green box means an object bounding box predicted from the anchor in the red/green patch~(if they coincide, we only show the red patches and boxes).}
		\label{poor-quality prediction}
		\vspace{-2mm}
\end{figure}

	Recent one-stage object detectors attempted to predict consistent outputs of the two separate tasks, by focusing on the center of an object~\cite{duan2019centernet,kong2020foveabox,tian2019fcos,zhang2020bridging}. They assume that an anchor~(\ie, an anchor-point for an anchor-free detector, or an anchor-box for an anchor-based detector) at the center of the object is likely to give more accurate predictions for both classification and localization. For example, recent FCOS~\cite{tian2019fcos} and ATSS~\cite{zhang2020bridging} both use a centerness branch to enhance classification scores predicted from the anchors near the center of the object, and assign larger weights to the localization loss for the corresponding anchors. Besides, FoveaBox~\cite{kong2020foveabox} regards the anchors inside a predefined central region of the object as positive samples.
	Such heuristic designs have achieved excellent results, but these methods might suffer from two limitations: 
	
\vspace{-3mm}
\paragraph{(1)~Independence of classification and localization.}
Recent one-stage detectors perform object classification and localization independently by using two separate branches in parallel (\ie,  heads). Such a two-branch design might cause a lack of interaction between the two tasks, leading to an inconsistency in predictions when performing them. As shown in the `Result' column in Figure~\ref{poor-quality prediction},
	an ATSS detector recognizes an object of `Dining table' (indicated by the anchor shown with a red patch), but localizes another object of `Pizza' more accurately~(red bounding box).
	
\vspace{-3mm}
\paragraph{(2)~Task-agnostic sample assignment.}
Most anchor-free detectors use a geometry-based assignment scheme to select anchor-points near the center of an object for both classification and localization~\cite{duan2019centernet,kong2020foveabox,zhang2020bridging}, while anchor-based detectors often assign anchor-boxes by computing IoUs between the anchor boxes and ground truth ~\cite{redmon2016you,ren2015faster,zhang2020bridging}. However, the optimal anchors for classification and localization are often inconsistent, and may vary considerably depending on the shape and characteristics of the objects. The widely used sample assignment scheme is task agnostic, and thus may be difficult to make an accurate yet consistent prediction for the two tasks, as demonstrated in `Score' and `IoU' distributions of ATSS in Figure~\ref{poor-quality prediction}. The `Result' column also illustrates that a spatial location of the best localization anchor~(green patch) can be not at the center of the object, and it is not well aligned with the best classification anchor~(red patch). As a result, a precise bounding box may be suppressed by the less accurate one during Non-Maximum Suppression~(NMS).

	\vspace{1mm}
	To address such limitations, we propose a Task-aligned One-stage Object Detection (TOOD) that aims to align the two tasks more accurately by designing a new head structure with an alignment-oriented learning approach:

\vspace{-3mm}
\paragraph{Task-aligned head.}
In contrast to the conventional head in one-stage object detection where classification and localization are implemented separately by using two branches in parallel, we design a Task-aligned head~(T-head) to enhance an \textit{interaction} between the two tasks. This allows the two tasks to work more collaboratively, which in turn aligns their predictions more accurately.
	\head is conceptually simple: it computes task-interactive features, and makes predictions via a novel Task-Aligned Predictor (TAP).
	Then it aligns spatial distributions of the two predictions according to the learning signals provided by a task alignment learning, as described next.
	
\vspace{-2mm}
\paragraph{Task alignment learning.}
    To further overcome the misalignment problem, we propose a Task Alignment Learning (TAL) to \textit{explicitly} pull closer the optimal anchors for the two tasks. It is performed by designing a sample assignment scheme and a task-aligned loss.
    The sample assignment collects training samples (\ie, positives or negatives) by computing a degree of task-alignment at each anchor, whereas the task-aligned loss gradually unifies the best anchors for predicting both classification and localization during the training.
    Therefore, at inference, a bounding box with the highest classification score and jointly having the most precise localization can be preserved.

The proposed T-head and learning strategy can work collaboratively towards making predictions with high quality in both classification and localization.
The main contributions of this work can be summarized as follows:
	(1)~we design a new T-head to enhance the interaction between classification and localization while maintaining their characteristics, and further align the two tasks at the predictions;
	(2)~we propose TAL to explicitly align the two tasks at the identified task-aligned anchors, as well as providing learning signals for the proposed predictor;
	(3)~we conducted extensive experiments on MSCOCO~\cite{lin2014microsoft}, where our TOOD achieved a \textbf{51.1 AP}, surpassing recent one-stage detectors such as ATSS~\cite{zhang2020bridging}, GFL~\cite{li2020generalized} and PAA~\cite{kim2020probabilistic}, by a large margin. Qualitative results further validate the effectiveness of our task-alignment approaches.

\vspace{-1mm}
\section{Related Work} \label{related-work}
\vspace{-1mm}
	\paragraph{One-stage detectors.}
	OverFeat~\cite{sermanet2013overfeat} is one of the earliest CNN-based one-stage detectors. Afterward, YOLO~\cite{redmon2016you} was developed to directly predict bounding boxes and classification scores, without an additional stage to generate region proposals. SSD~\cite{liu2016ssd} introduces anchors with multi-scale predictions from multi-layer convolutional features, and  Focal loss~\cite{lin2017focal} was proposed to address the problem of class imbalance for one-stage detectors like RetinaNet. Keypoint-based detection methods, such as~\cite{duan2019centernet,law2018cornernet,zhou2019objects}, address the detection problem by identifying and grouping multiple key points of a bounding box. Recently, FCOS~\cite{tian2019fcos} and FoveaBox~\cite{kong2020foveabox} were developed to locate objects of interest via anchor-points and point-to-boundary distances. Most mainstream one-stage detectors are composed of two FCN-based branches for classification and localization, which may lead to the misalignment between the two tasks. In this paper, we enhance the alignment between the two tasks via a new head structure and an alignment-oriented learning approach.

\vspace{-3mm}
	\paragraph{Training sample assignment.}
	Most anchor-based detectors such as ~\cite{redmon2016you,zhang2020bridging}, collect training samples by computing IoUs between proposals and ground truth, while an anchor-free detector regards the anchors inside the center region of an object as positive samples ~\cite{duan2019centernet,kong2020foveabox,tian2019fcos}.
    Recent studies attempted to train the detectors more effectively by collecting more informative training samples using output results.
	For example, FSAF~\cite{zhu2019feature} selects meaningful samples from feature pyramids based on the computed loss, and similarly, SAPD~\cite{zhu2020soft} provides a soft-selection version of FSAF by designing a meta-selection network.
	FreeAnchor~\cite{zhang2019freeanchor} and MAL~\cite{ke2020multiple} identify the best anchor-box by computing the losses in an effort to improve the matching between anchors and objects. PAA~\cite{kim2020probabilistic} adaptively separates the anchors into positive and negative samples by fitting a probability distribution to the anchor scores. 
	Mutual Guidance~\cite{zhang2020localize} improves anchor assignment for one task by considering the prediction quality on the other task.
	Different from the positive/negative sample assignment, PISA~\cite{cao2020prime} re-weights the training samples according to the precision rank of the outputs.
	Noisy Anchor~\cite{li2020learning} assigns soft labels to the training samples, and re-weights the anchor-boxes using a cleanliness score to mitigate the noise incurred by binary labels.
	GFL~\cite{li2020generalized} replaces the binary classification label with an IoU score to integrate the localization quality into classification.
	These excellent approaches inspired the current work to develop a new assignment mechanism from a task-alignment point of view.

\begin{figure}[t]
		\centering
		\includegraphics[width=8cm]{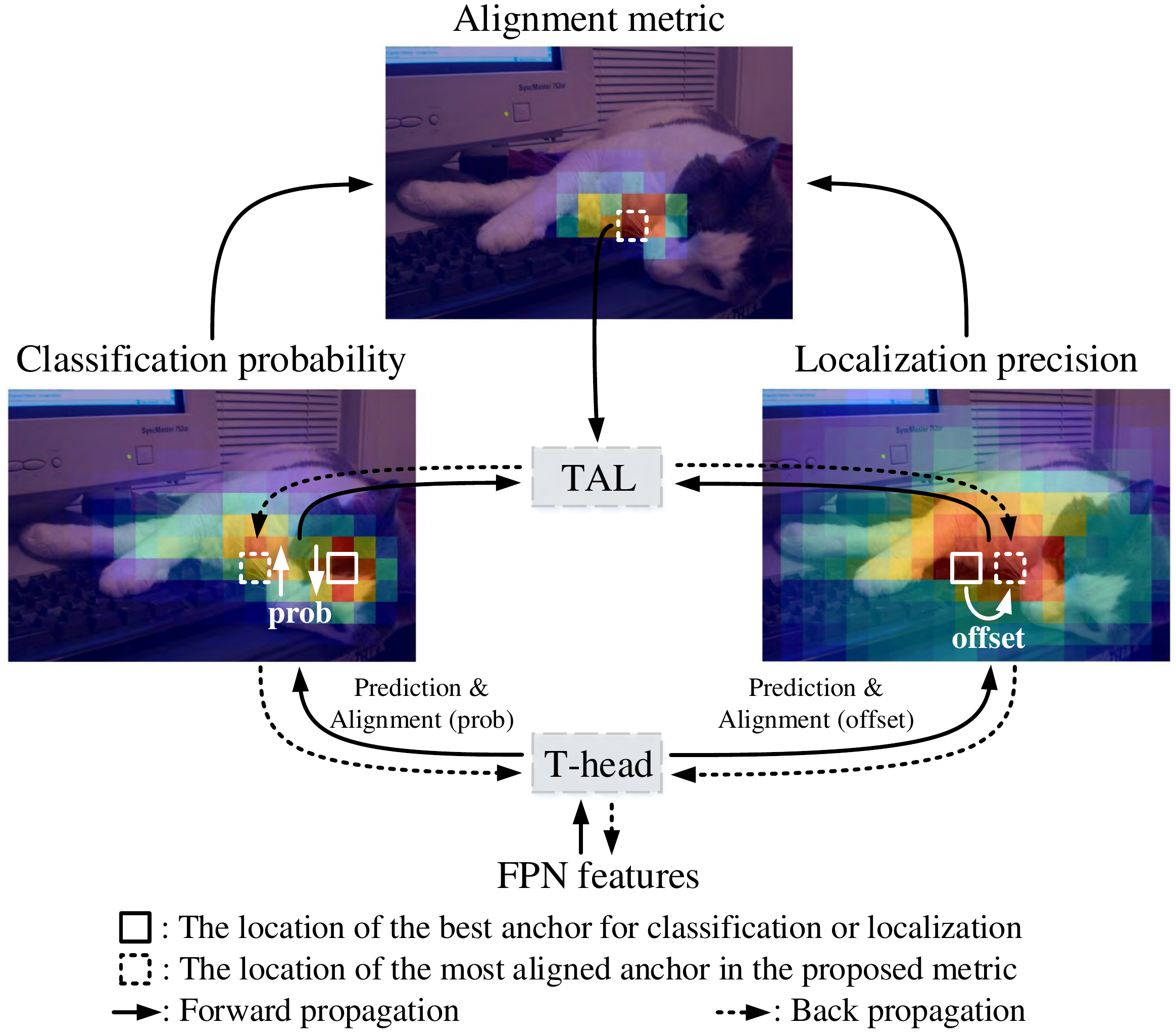}
		\caption{Overall learning mechanism of TOOD. 
		First, predictions are made by \head on the FPN features. 
		Second, the predictions are used to compute a task alignment metric at each anchor point, based on which TAL produces learning signals for \head.
		Lastly, \head adjusts the distributions of classification and localization accordingly.
		Specifically, the most aligned anchor obtains a higher classification score through  `prob'~(probability map), and acquires a more accurate bounding box prediction via a learned `offset'.
		}
		\label{tood_demo}
		\vspace{-2.5mm}
\end{figure}

\section{Task-aligned One-stage Object Detection}
\vspace{-0.5mm}
     \paragraph{Overview.}
	Similar to recent one-stage detectors such as~\cite{li2020generalized,zhang2020bridging}, the proposed \name has an overall pipeline of `backbone-FPN-head'.
	Moreover, by considering efficiency and simplicity, \name uses a single anchor per location (same as ATSS~\cite{zhang2020bridging}), where the `anchor' means an anchor point for an anchor-free detector, or an anchor box for an anchor-based detector.
	As discussed, existing one-stage detectors have limitations of task misalignment between classification and localization, due to the divergence of two tasks which are often implemented using two separate head branches.
	In this work, we propose to align the two tasks more explicitly using a designed Task-aligned head~(T-head) with a new Task Alignment Learning~(TAL).
	As illustrated in Figure~\ref{tood_demo}, \head and TAL can work collaboratively to improve the alignment of two tasks. 
	Specifically, \head first makes predictions for the classification and localization on the FPN features. Then TAL computes task alignment signals based on a new task alignment metric which measures the degree of alignment between the two predictions.
	Lastly, \head automatically adjusts its classification probabilities and localization predictions using learning signals computed from TAL during back propagation. 
	
\begin{figure*}[tb]
		\centering
		\subfigure[Parallel head]{
			\includegraphics[height=3.7cm]{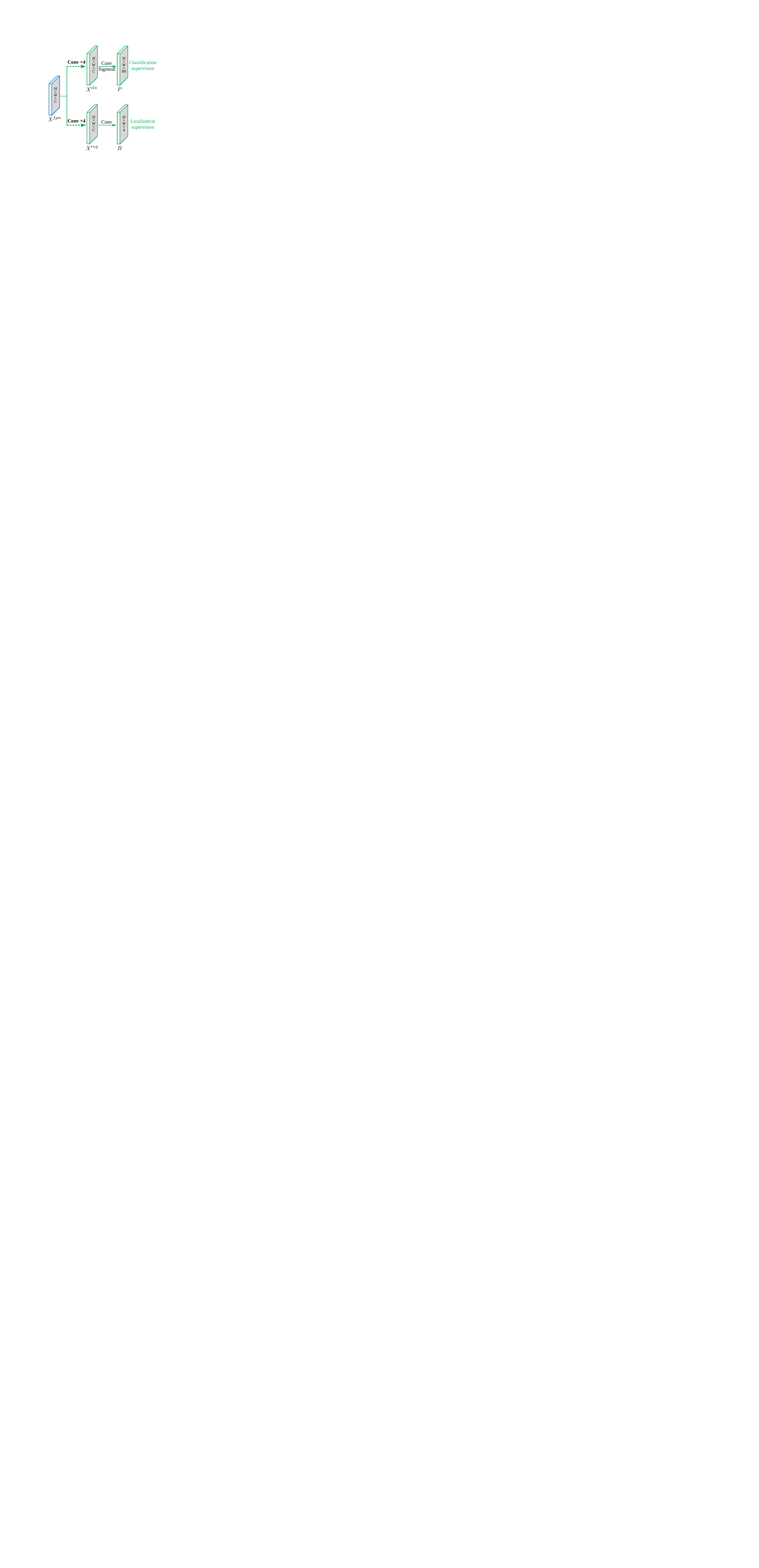}
			\label{ph}
		}
		\ \ \ 
		\subfigure[Task-aligned head (T-Head)]{
			\includegraphics[height=3.7cm]{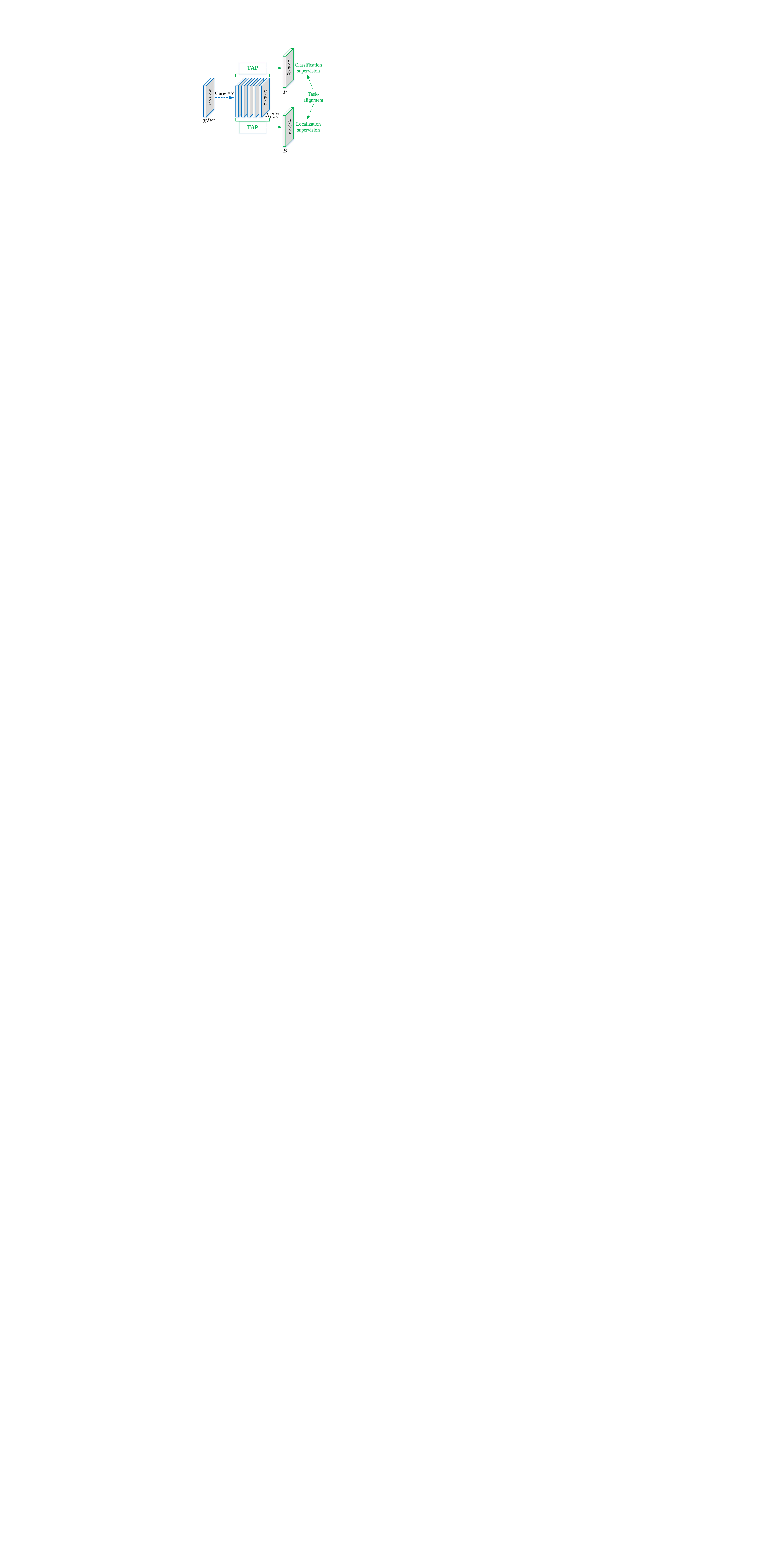}
			\label{tah}
		}
		\ \ \ 
        \subfigure[Task-aligned predictor (TAP)]{
			\includegraphics[height=3.7cm]{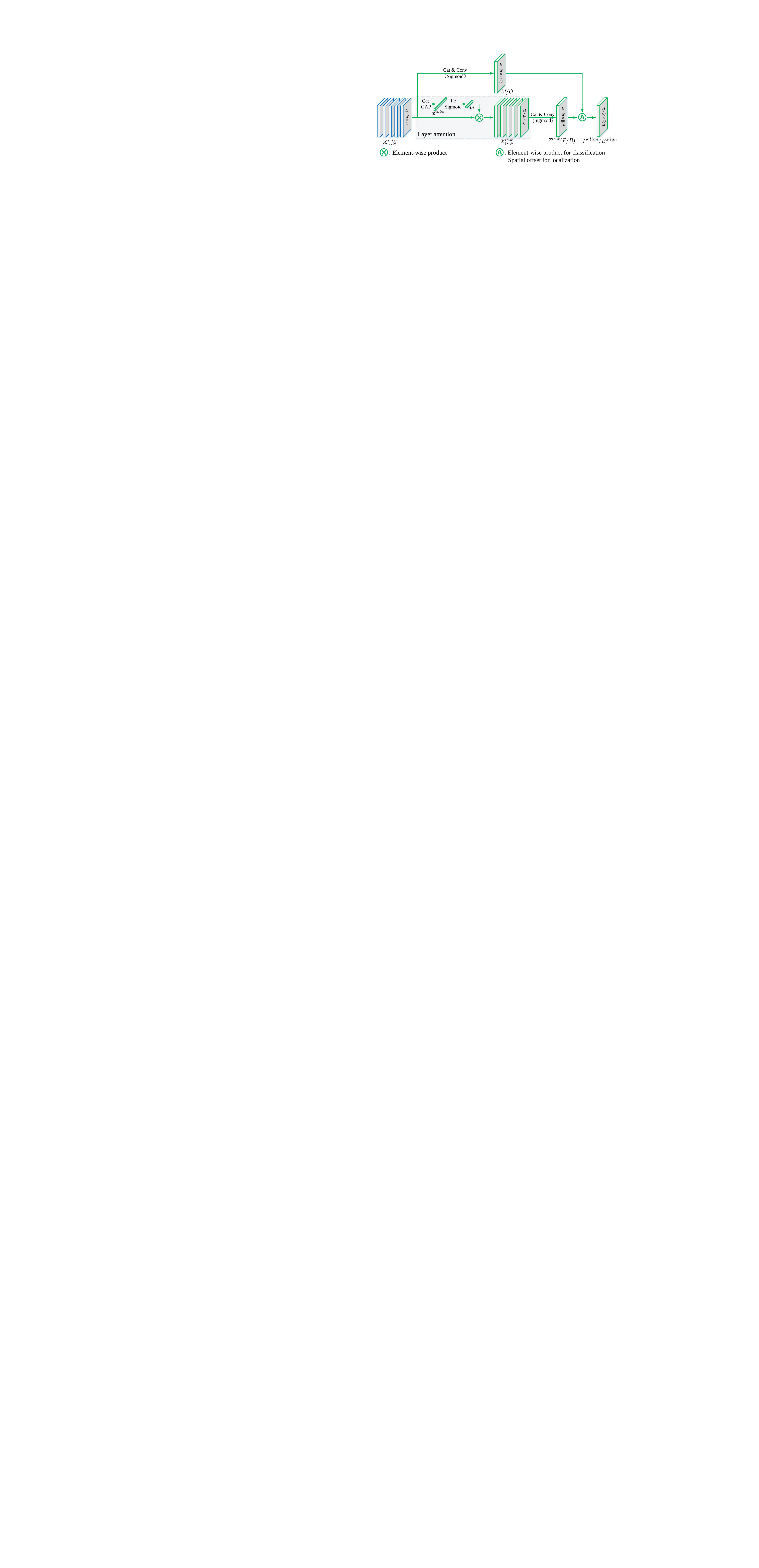}
			\label{tap}
		}
        \caption{Comparison between the conventional parallel head and the proposed T-Head.}
		\label{network architectures}
\end{figure*}	
	
\subsection{Task-aligned Head} \label{feature-alignment}
    Our goal is to design an efficient head structure to improve the conventional design of the head in one-stage detectors (as shown in Figure~\ref{ph}). In this work, we achieve this by considering two aspects: (1) increasing the interaction between the two tasks, and (2) enhancing the detector ability of learning the alignment.
	The proposed T-head is shown in Figure~\ref{tah},
	where it has a simple feature extractor with two Task-Aligned Predictors~(TAP).

	To enhance the interaction between classification and localization, we use a feature extractor to learn a stack of \textit{task-interactive} features from multiple convolutional layers, as shown by the blue part in Figure~\ref{tah}. This design not only facilitates the task interaction, but also provides multi-level features with multi-scale effective receptive fields for the two tasks. Formally, let $X^{fpn} \in \mathbb{R}^{H \times W \times C}$ denotes the FPN features, where $H$, $W$ and $C$ indicate height, width and the number of channels, respectively. The feature extractor uses $N$ consecutive conv layers with activation functions to compute the \textit{task-interactive} features:
	\begin{equation}
	\resizebox{.90\linewidth}{!}{$
    \displaystyle
	X^{inter}_{k}=\left\{
		\begin{aligned}
		&\delta(conv_{k}(X^{fpn})), k=1 \\
		&\delta(conv_{k}(X^{inter}_{k-1})), k>1
	\end{aligned}
	\right., {\forall} k \in \{1,2,...,N\},
	$}
	\end{equation}
	where $conv_{k}$ and $\delta$ refer to the $k$-th conv layer and a $relu$ function, respectively.
Thus we extract rich multi-scale features from the FPN features using a single branch in the head.
Then, the computed task-interactive features will be fed into two TAP for aligning classification and localization.

\vspace{-2mm}
\paragraph{Task-aligned Predictor (TAP).}
    We perform both object classification and localization on the computed task-interactive features, where the two tasks can well perceive the state of each other.
    However, due to the single branch design, the task-interactive features inevitably introduce a certain level of feature conflicts between two different tasks, which have also been discussed in~\cite{song2020revisiting,wu2019double}. Intuitively, the tasks of object classification and localization have different targets, and thus focus on different types of features (\eg, different levels or receptive fields).
	Consequently, we propose a layer attention mechanism to encourage task decomposition by dynamically computing such \textit{task-specific} features at the layer level.
	As shown in Figure~\ref{tap}, the task-specific features are computed separately for each task of classification or localization:
	\begin{equation}
	X^{task}_{k} = \bm{w}_{k} \cdot X^{inter}_{k}, {\forall} k \in \{1,2,...,N\},
	\end{equation}
	where $\bm{w}_{k}$ is the $k$-th element of the  learned layer attention $\bm{w} \in \mathbb{R}^{N}$. $\bm{w}$ is computed from the cross-layer task-interactive features, and is able to capture the dependencies between layers:
	\begin{equation}
	\bm{w} = \sigma(fc_{2}(\delta(fc_{1}(\bm{x}^{inter})))),
	\end{equation}
	where $fc_{1}$ and $fc_{2}$ refer to two fully-connected layers. $\sigma$ is a $sigmoid$ function, and $\bm{x}^{inter}$ is obtained by applying an average pooling to $X^{inter}$ which are the concatenated features of $X^{inter}_{k}$.
	Finally, the results of classification or localization are predicted from each $X^{task}$:
	\begin{equation}
	Z^{task} = conv_{2}(\delta(conv_{1}(X^{task}))),
	\end{equation}
	where  $X^{task}$ is the concatenated features of $X^{task}_{k}$, and $conv_{1}$ is a 1$\times$1 conv layer for dimension reduction. $Z^{task}$ is then converted into dense classification scores $P \in \mathbb{R}^{H \times W \times 80}$ using $sigmoid$ function, or object bounding boxes $B \in \mathbb{R}^{H \times W \times 4}$ with $distance$-$to$-$bbox$ conversion as applied in~\cite{tian2019fcos,zhang2020bridging}.

\vspace{-2mm}
\paragraph{Prediction alignment.}
At the prediction step, we further align the two tasks explicitly by adjusting the spatial distributions of the two predictions: $P$ and $B$.
Different from the previous works using a centerness branch~\cite{tian2019fcos} or an IoU branch~\cite{kim2020probabilistic} which can only adjust the classification prediction based on either classification features or localization features, we align the two predictions by considering both tasks jointly using the computed task-interactive features.
Notably, we perform the alignment method separately on the two tasks.
As shown in Figure~\ref{tap}, we use a spatial probability map $M \in \mathbb{R}^{H \times W \times 1}$ to adjust the classification prediction:
	\begin{equation}
	P^{align} = \sqrt{P \times M},
	\end{equation}
where $M$ is computed from the interactive features, allowing it to learn a degree of consistency between the two tasks at each spatial location. 

Meanwhile, to make an alignment on localization prediction, we further learn spatial offset maps $O \in \mathbb{R}^{H \times W \times 8}$ from the interactive features, which are used to adjust the predicted bounding box  at each location. Specifically, the learned spatial offset enables the most aligned anchor point to identify the best boundary predictions around it:
	\begin{equation}
	\resizebox{.90\linewidth}{!}{$
	B^{align}(i, j, c) = \\
	B(i + O(i, j, 2 \times c), j + O(i, j, 2 \times c + 1), c),
	\label{loc-align}
	$}
	\end{equation}
where an index~$(i, j, c)$ denotes the $(i, j)$-th spatial location at the $c$-th channel in a tensor. Eq.(\ref{loc-align}) is implemented by bilinear interpolation, and its computational overhead is negligible due to the very small channel dimension of $B$. Noteworthily, offsets are learned independently for each channel, which means each boundary of the object has its own learned offset.
This allows for a more accurate prediction of the four boundaries because each of them can individually learn from the most precise anchor point near it.
Therefore, our method not only aligns the two tasks, but also improves the localization accuracy by identifying a precise anchor point for each side.

The alignment maps $M$ and $O$ are learned automatically from the stack of interactive features:
	\begin{equation}
	M = \sigma(conv_{2}(\delta(conv_{1}(X^{inter}))))
	\end{equation}
	\begin{equation}
	O = conv_{4}(\delta(conv_{3}(X^{inter})))
	\end{equation}
	where $conv_{1}$ and $conv_{3}$ are two 1$\times$1 conv layers for dimension reduction.
	The learning of $M$ and $O$ is performed by using the proposed Task Alignment Learning~(TAL) which will be described next.
    Notice that our \head is an independent module and can work well without TAL. It can be readily applied to various one-stage object detectors in a plug-and-play manner to improve detection performance.

\subsection{Task Alignment Learning}
	\label{anchor-point-alignment}
    We further introduce a Task Alignment Learning~(TAL) that further guides our \head to make task-aligned predictions.
	TAL differs from previous methods~\cite{cao2020prime,ke2020multiple,kim2020probabilistic,li2020learning,li2020generalized,zhang2020localize,zhang2019freeanchor} in two aspects. 
	First, from the task-alignment point of view, it dynamically selects high-quality anchors based on a designed metric.
	Second, it considers both anchor assignment and weighting simultaneously.
	It comprises a sample assignment strategy and new losses designed specifically for aligning the two tasks.
	
\vspace{-2mm}
	\subsubsection{Task-aligned Sample Assignment}
	\label{subsubsec:tsa}
\vspace{-1.5mm}
	To cope with NMS, the anchor assignment for a training instance should satisfy the following rules:
	(1)~a well-aligned anchor should be able to predict a high classification score with a precise localization jointly;
	(2)~a misaligned anchor should have a low classification score and be suppressed subsequently.
	With the two objectives, we design a new anchor alignment metric to explicitly measure the degree of task-alignment at the anchor level. The alignment metric is integrated into the sample assignment and loss functions to dynamically refine the predictions at each anchor.

\vspace{-3mm}
	\paragraph{Anchor alignment metric.}
	Considering that a classification score and an IoU between the predicted bounding box and the ground truth indicate the quality of the predictions by the two tasks,
	we measure the degree of task-alignment using a high-order combination of the classification score and the IoU.
	To be specific, we design the following metric to compute anchor-level alignment for each instance:
	\begin{equation}
	t = s^{\alpha} \times u^{\beta},
	\label{eq-metric}
	\end{equation}
	where $s$ and $u$ denote a classification score and an IoU value, respectively. $\alpha$ and $\beta$ are used to control the impact of the two tasks in the anchor alignment metric.
	Notably, $t$ plays a critical role in the joint optimization of the two tasks towards the goal of task-alignment. It encourages the networks to dynamically focus on high-quality (\ie, task-aligned) anchors from the perspective of joint optimization.

\vspace{-3mm}
	\paragraph{Training sample assignment.} As discussed in~\cite{zhang2020bridging,zhang2019freeanchor}, training sample assignment is crucial to the training of object detectors.
	To improve the alignment of two tasks, we focus on the task-aligned anchors, and adopt a simple assignment rule to select the training samples: for each instance, we select $m$ anchors having the largest $t$ values as positive samples, while using the remaining anchors as negative ones.
	Again, the training is performed by computing new loss functions designed specifically for aligning the tasks of classification and localization.
	
\begin{table*}
\small
\centering
\begin{tabular}{lrrrrrr}
\toprule
Method & Head & Head/full Params~(M) & Head/full FLOPs~(G) & AP & AP$_{50}$ & AP$_{75}$ \\
\cmidrule(r){1-1}
\cmidrule(r){2-2}
\cmidrule(r){3-4}
\cmidrule(r){5-7}
FoveaBox~\cite{kong2020foveabox}            & Parallel head & 4.92/36.20 & 104.87/206.28 & 37.3 & 56.2 & 39.7
\\
					& \textbf{\head} & 4.82/36.10  & 100.79/202.20 & \textbf{38.0} & \textbf{56.8} & \textbf{40.5}
					\\
\cmidrule(r){1-1}
\cmidrule(r){2-2}
\cmidrule(r){3-4}
\cmidrule(r){5-7}
FCOS w/ imprv~\cite{tian2019fcos}                & Parallel head & 4.92/32.02 & 104.91/200.50 & 38.6 & 57.2 & 41.7    \\
					& \textbf{\head} & 4.82/31.92 & 100.79/196.38 & \textbf{40.5} & \textbf{58.5} & \textbf{43.8}  \\
\cmidrule(r){1-1}
\cmidrule(r){2-2}
\cmidrule(r){3-4}
\cmidrule(r){5-7}
ATSS~(anchor-based)~\cite{zhang2020bridging}                & Parallel head & 4.92/32.07 & 104.87/205.21 & 39.3 & 57.5 & 42.8     \\
					& \textbf{\head} & 4.82/31.98 & 100.79/201.13 & \textbf{41.1} & \textbf{58.6} & \textbf{44.5}      \\
\cmidrule(r){1-1}
\cmidrule(r){2-2}
\cmidrule(r){3-4}
\cmidrule(r){5-7}
ATSS~(anchor-free)~\cite{zhang2020bridging}                & Parallel head & 4.92/32.07 & 104.87/205.21 & 39.2 & 57.4 & 42.2     \\
					& \textbf{\head} & 4.82/31.98 & 100.79/201.13 & \textbf{41.1} & \textbf{58.4} & \textbf{44.5}      \\
\bottomrule
\end{tabular}
\caption{Comparisons between different head structures in various detectors. FLOPs are measured on the input image size of 1280$\times$800.}
\label{ex-head-structure}
\vspace{-2mm}
\end{table*}	
	
\subsubsection{Task-aligned Loss}
	\label{subsubsec:loss}
	\paragraph{Classification objective.}
	To explicitly increase classification scores for the aligned anchors, and at the same time, reduce the scores of the misaligned ones (\ie, having a small $t$), we use $t$ to replace the binary label of a positive anchor during training.
	However, we found that the network cannot converge when the labels~(\ie, $t$) of the positive anchors become small with the increase of $\alpha$ and $\beta$. Therefore, we use a normalized $t$, namely $\hat{t}$, to replace the binary label of the positive anchor,
	where $\hat{t}$ is normalized by the following two properties: (1)~to ensure effective learning of hard instances~(which usually have a small $t$ for all corresponding positive anchors); (2)~to preserve the rank between instances based on the precision of the predicted bounding boxes. Thus, we adopt a simple instance-level normalization to adjust the scale of $\hat{t}$: the maximum of $\hat{t}$ is equal to the largest IoU value ($u$) within each instance.
	Then $Binary$ $Cross$ $Entropy$~($BCE$) computed on the positive anchors  for the classification task can be rewritten as,
	\begin{equation}
	L_{cls\_pos} = \sum_{i = 1}^{N_{pos}} BCE(s_{i}, \hat{t}_{i}),
	\label{eq:cls_pos}
	\end{equation}
	where $i$ denotes the $i$-th anchor from the $N_{pos}$ positive anchors corresponding to one instance.
	Following~\cite{lin2017focal}, we employ a focal loss for classification to mitigate the imbalance between the negative and positive samples during training.
	The focal loss computed on the positive anchors can be reformulated by Eq.(\ref{eq:cls_pos}), and the final loss function for the classification task is defined as follows:
	\begin{equation}
	\resizebox{.85\linewidth}{!}{$
    \displaystyle
	L_{cls} = \sum_{i = 1}^{N_{pos}} \left| \hat{t}_{i} - s_{i} \right|^{\gamma} BCE(s_{i}, \hat{t}_{i}) + \sum_{j = 1}^{N_{neg}} s_{j}^{\,\,\gamma} BCE(s_{j}, 0),
	$}
	\label{eq-revised-focal-loss}
	\end{equation}
	where $j$ denotes the $j$-th anchor from the $N_{neg}$ negative anchors, and $\gamma$ is the focusing parameter~\cite{lin2017focal}.
	
	\vspace{-2mm}
	\paragraph{Localization objective.}
	A bounding box predicted by a well-aligned anchor~(\ie, having a large $t$) usually has both a large classification score with a precise localization, and such a bounding box is more likely to be preserved during NMS. In addition, $t$ can be applied for selecting high-quality bounding boxes by weighting the loss more carefully to improve the training. As discussed in~\cite{pang2019libra}, learning from high-quality bounding boxes is beneficial to the performance of a model, while the low-quality ones often have a negative impact on the training by producing a large amount of less informative and redundant signals to update the model. In our case, we apply the $t$ value for measuring the quality of a bounding box.
	Thus, we improve the task alignment and regression precision by focusing on the well-aligned anchors (with a large $t$), while reducing the impact of the misaligned anchors (with a small $t$) during bounding box regression.
	Similar to the classification objective, we re-weight the loss of bounding box regression computed for each anchor based on $\hat{t}$, and a $GIoU$ loss~($L_{GIoU}$)~\cite{rezatofighi2019generalized} can be reformulated as follows:
	\begin{equation}
	L_{reg} = \sum_{i = 1}^{N_{pos}} \hat{t}_{i} L_{GIoU}(b_{i}, \bar{b_{i}}),
	\label{eq-smooth-l1-loss}\
	\end{equation}
	where $b$ and $\bar{b}$ denote the predicted bounding boxes and the corresponding ground-truth boxes.
	The total training loss for TAL is the sum of $L_{cls}$ and $L_{reg}$.
	
\vspace{-1mm}
\section{Experiments and Results}
	\paragraph{Dataset and evaluation protocol.}
	All experiments are implemented on the large-scale detection benchmark MS-COCO $2017$~\cite{lin2014microsoft}. Following the standard practice~\cite{lin2017feature,lin2017focal}, we use the $trainval135k$ set~(115$K$ images) for training and $minival$ set~(5$K$ images) as validation for our ablation study. We report our main results on the $test$-$dev$ set for comparison with the state-of-the-art detectors. The performance is measured by COCO Average Precision~(AP)~\cite{lin2014microsoft}.
	
	\vspace{-2mm}
	\paragraph{Implementation details.} \label{implementation-details}
	As with most one-stage detectors~\cite{kong2020foveabox,lin2017focal,tian2019fcos}, we use the detection pipeline of `backbone-FPN-head', with different backbones including ResNet-50, ResNet-101 and ResNeXt-101-64$\times$4d pre-trained on ImageNet~\cite{deng2009imagenet}.
	Similar to ATSS~\cite{zhang2020bridging}, TOOD tiles one anchor per location. Unless specified, we report experimental results of an anchor-free TOOD~(an anchor-based TOOD can achieve a similar performance as shown in Table~\ref{ex-tuh-tsa}).
	The number of interactive layers $N$ is set as 6 to make \head have a similar number of parameters as the conventional parallel head, and the focusing parameter $\gamma$ is set to 2 as used in~\cite{li2020generalized,lin2017focal}.
	More implementation and training details are presented in Supplementary Material (SM).

\subsection{Ablation Study}
\begin{figure*}[ht]
		\centering
		\small
		\setlength{\tabcolsep}{0.5mm}{
			\renewcommand\arraystretch{0.55}
			\begin{tabular}{ccccccc}
			& \multirow{6}*{\includegraphics[height=1.75cm]{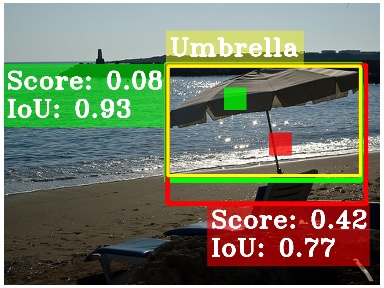}}
			& \multirow{6}*{\includegraphics[height=1.75cm]{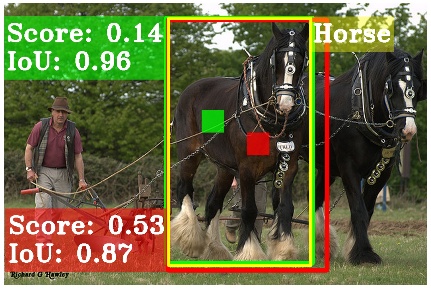}}
			& \multirow{6}*{\includegraphics[height=1.75cm]{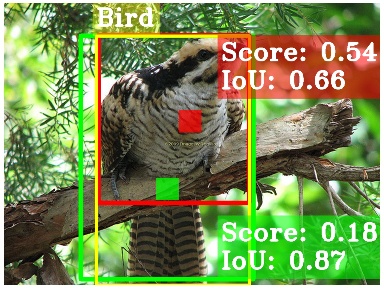}}
			& \multirow{6}*{\includegraphics[height=1.75cm]{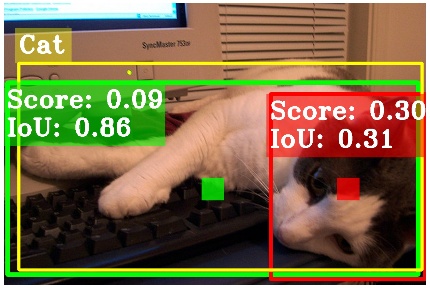}}
			& \multirow{6}*{\includegraphics[height=1.75cm]{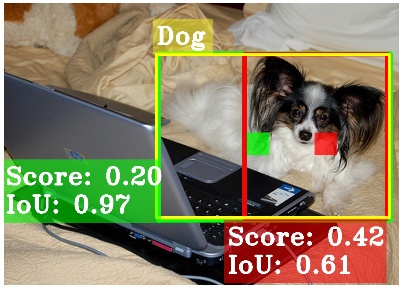}}
			& \multirow{6}*{\includegraphics[height=1.75cm]{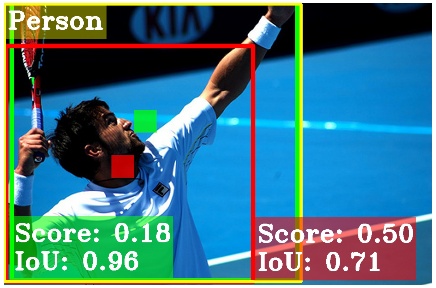}}
			~ \\~ \\ Parallel head \\ + \\ ATSS \\ ~ \\~ \\ ~ \\
			
			& \multirow{6}*{\includegraphics[height=1.75cm]{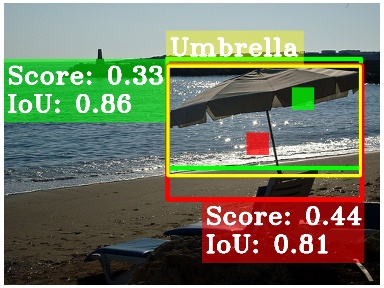}}
			& \multirow{6}*{\includegraphics[height=1.75cm]{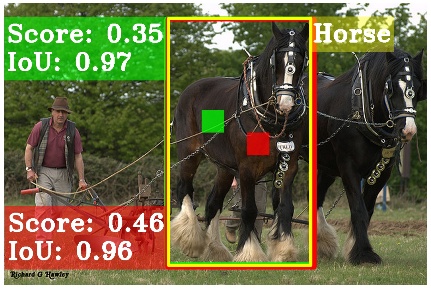}}
			& \multirow{6}*{\includegraphics[height=1.75cm]{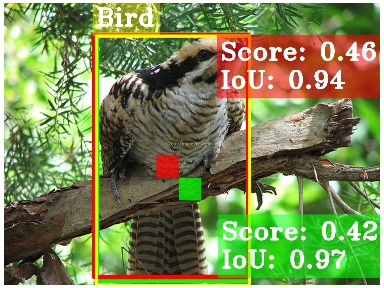}}
			& \multirow{6}*{\includegraphics[height=1.75cm]{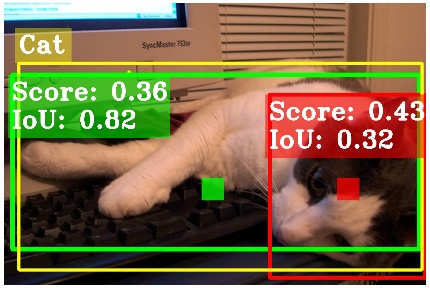}}
			& \multirow{6}*{\includegraphics[height=1.75cm]{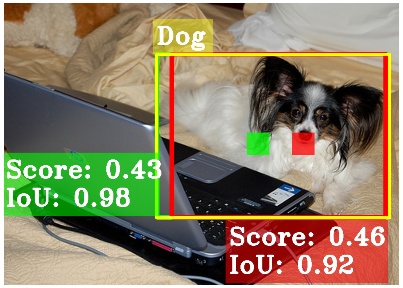}}
			& \multirow{6}*{\includegraphics[height=1.75cm]{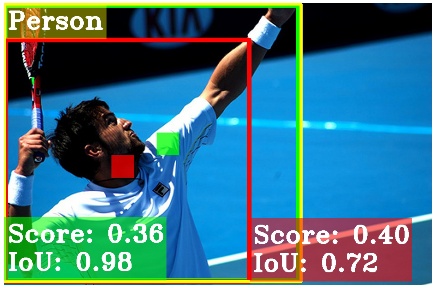}}
			~ \\~ \\ \textbf{\head} \\ + \\ ATSS \\ ~ \\~ \\ ~ \\
			
			& \multirow{6}*{\includegraphics[height=1.7cm]{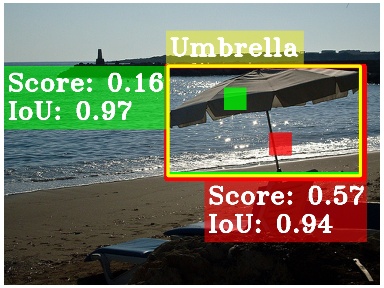}}
			& \multirow{6}*{\includegraphics[height=1.75cm]{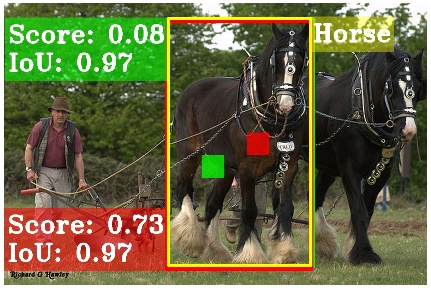}}
			& \multirow{6}*{\includegraphics[height=1.75cm]{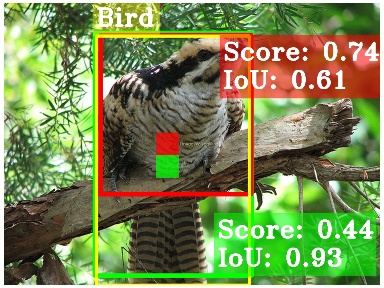}}
			& \multirow{6}*{\includegraphics[height=1.75cm]{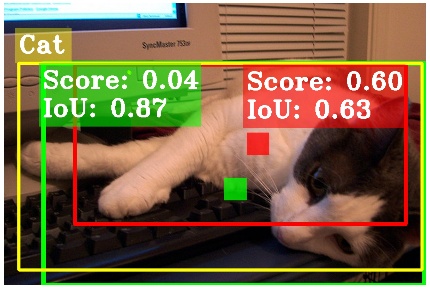}}
			& \multirow{6}*{\includegraphics[height=1.75cm]{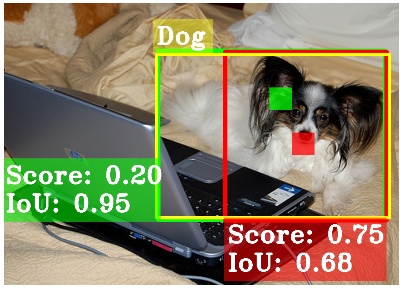}}
			& \multirow{6}*{\includegraphics[height=1.75cm]{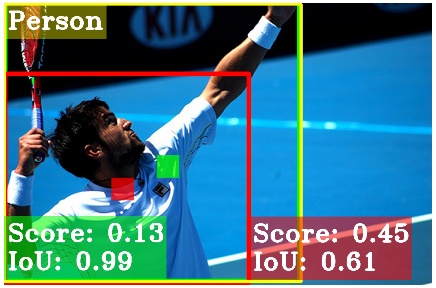}}
			~ \\~ \\ Parallel head \\ + \\ \textbf{TAL} \\ ~ \\~ \\ ~ \\
			
			& \multirow{6}*{\includegraphics[height=1.75cm]{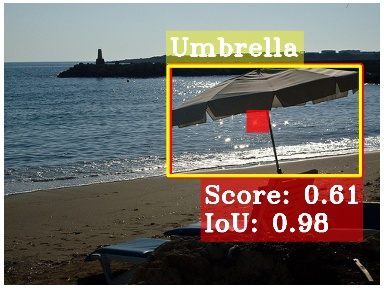}}
			& \multirow{6}*{\includegraphics[height=1.75cm]{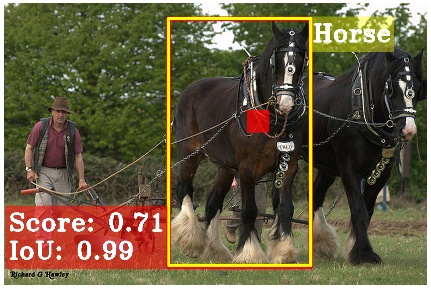}}
			& \multirow{6}*{\includegraphics[height=1.75cm]{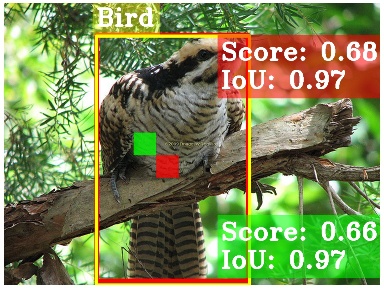}}
			& \multirow{6}*{\includegraphics[height=1.75cm]{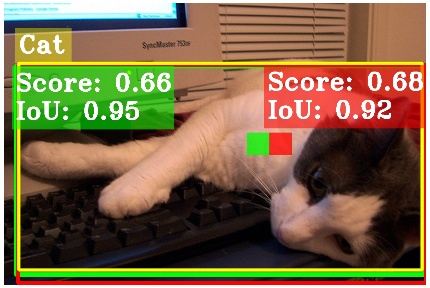}}
			& \multirow{6}*{\includegraphics[height=1.75cm]{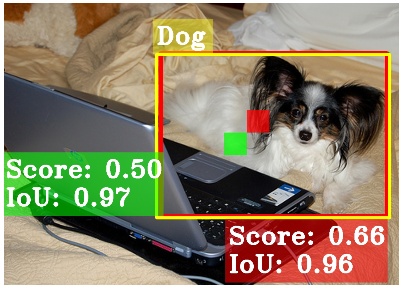}}
			& \multirow{6}*{\includegraphics[height=1.75cm]{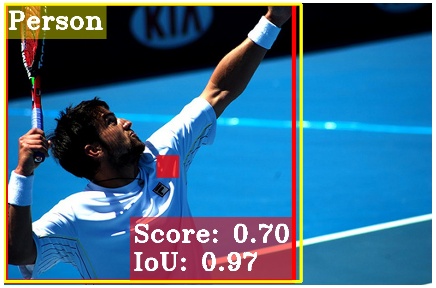}}
			~ \\~ \\ \textbf{\head} \\ + \\ \textbf{TAL} \\ ~ \\~ \\ ~ \\
			
			\end{tabular}
		}
		\caption{Illustration of several detection results predicted from the best anchors for classification~(in red) and localization~(in green). The illustrated patches and bounding boxes correspond to that in Figure~\ref{poor-quality prediction}.}
		\label{illustration_of_images}
		\vspace{-3mm}
\end{figure*}

	For an ablation study, we use the ResNet-50 backbone and train the model for 12 epochs unless specified. The performances are reported on COCO $minival$ set.
	
\vspace{-3mm}
	\paragraph{On head structures.}
	We compare our \head with the conventional parallel head in Table~\ref{ex-head-structure}. It can be adopted in various one-stage detectors in a plug-and-play manner, and consistently outperforms the conventional head by 0.7 to 1.9 AP, with fewer parameters and FLOPs. This validates the effectiveness of our design, and demonstrates that \head can work more efficiently with higher performance, by introducing task interaction and prediction alignment.

\begin{table}
\small
\centering
\begin{tabular}{lrrp{0.4cm}>{\centering}p{0.4cm}<{\centering}p{0.4cm}<{\centering}}
\toprule
Anchor assignment & Pos/neg & Weight & AP & AP$_{50}$ & AP$_{75}$ \\
\cmidrule(r){1-1}
\cmidrule(r){2-3}
\cmidrule(r){4-6}
IoU-based~\cite{lin2017focal}         & fixed & fixed & 36.5 & 55.5 & 38.7   \\
Center sampling~\cite{kong2020foveabox}   & fixed & fixed & 37.3 & 56.2 & 39.3   \\
Centerness~\cite{tian2019fcos}        & fixed & fixed & 37.4 & 56.1 & 40.3    \\
ATSS~\cite{zhang2020bridging} & fixed   & fixed & 39.2 & 57.4 & 42.2   \\
PISA~\cite{cao2020prime}              & fixed & ada   & 37.3 & 56.5 & 40.3   \\
NoisyAnchor~\cite{li2020learning}       & fixed & ada   & 38.0 & 56.9 & 40.6    \\
ATSS+QFL~\cite{li2020generalized} & fixed & ada & 39.9 & 58.5 & 43.0 \\
FreeAnchor~\cite{zhang2019freeanchor}        & ada   & fixed & 39.1 & 58.2 & 42.1    \\
MAL~\cite{ke2020multiple}               & ada   & fixed & 39.2 & 58.0 & 42.3   \\
PAA~\cite{kim2020probabilistic}$^{*}$               & ada   & fixed & 39.9 & 59.1 & 42.8    \\
PAA+IoU pred.~\cite{kim2020probabilistic}$^{*}$               & ada   & fixed & 40.9 & 59.4 & 43.9    \\
\textbf{TAL} & ada   & ada   & 40.3 & 58.5 & 43.8    \\
\textbf{TAL}$^{*}$ & ada   & ada   & 40.9 & 59.3 & 44.3    \\
\textbf{TAL} + \textbf{TAP}$^{*}$ & ada   & ada   & \textbf{42.5} & \textbf{60.3} & \textbf{46.4}    \\
\bottomrule
\end{tabular}
\caption{Comparisons between different schemes of training sample assignments. `Pos/neg': positive/negative anchor assignment. `Weight': anchor weight assignment. `fixed': fixed assignment. `ada': adaptive assignment. Here TAP aligns the predictions based on both classification and localization features from the last head tower. $^{*}$ indicates the model is trained for 18 epochs.}
\label{ex-sample-assignment}
\vspace{-1mm}
\end{table}

\vspace{-2mm}
	\paragraph{On sample assignments.} To demonstrate the effectiveness of TAL, we compare TAL with other learning methods using different sample assignment methods, as shown in Table~\ref{ex-sample-assignment}. 
	Training sample assignment can be divided into the fixed assignment and adaptive assignment according to whether it is a learning-based method.
	Different from the existing assignment methods, TAL adaptively assigns both positive and negative anchors, and at the same time, computes the weights of positive anchors more carefully, resulting in higher performance.
	To compare with PAA (+IoU pred.) which has an additional prediction structure, we integrate TAP into TAL, resulting in a higher AP of 42.5.
	More discussions on the differences between TAL and previous methods are presented in SM. 
	
	\vspace{-3mm}
	\paragraph{TOOD.} We evaluate the performance of the complete TOOD~(\head + TAL). As shown in Table~\ref{ex-tuh-tsa}, the anchor-free TOOD and anchor-based TOOD can achieve similar performance, \ie, 42.5 AP and 42.4 AP. Compared with ATSS, TOOD improves the performance of $\sim$3.2 AP.
	To be more specific, the improvements on AP$_{75}$ are significant, which yields $\sim$3.8 points higher AP in TOOD. This validates that aligning the two tasks can improve the detection performance.
	Notably, TOOD brings a higher improvement~(+3.3 AP) than the sum of the individual improvements by \head + ATSS~(+1.9 AP) and Parallel head + TAL~(+1.1 AP), as shown in Table~\ref{quantitative-analysis}. It suggests that \head and TAL can compensate strongly to each other.
	
	\vspace{-3mm}
	\paragraph{On hyper-parameters.}
	We first investigate the performance using different values of $\alpha$ and $\beta$ for TAL, which control the influence of classification confidence and localization precision on anchor alignment metric via $t$. Through a coarse search shown in Table~\ref{ex-hyper-parameters-t}, we adopt $\alpha=1$ and $\beta=6$ for our TAL.
	We then conduct several experiments to study the robustness of the hyper-parameter $m$, which is used to select positive anchors. 
	We use different values of $m$ in [5, 9, 13, 17, 21], and achieve results in a range of 42.0$\sim$42.5 AP, which suggests the performance is insensitive to $m$. Thus, we adopt $m=13$ in all our experiments.

\begin{table}
\centering
\small
\begin{tabular}{lrccc}
\toprule
Type & Method & AP & AP$_{50}$ & AP$_{75}$ \\
\cmidrule(r){1-1}
\cmidrule(r){2-2}
\cmidrule(r){3-5}
Anchor-free & ATSS~\cite{zhang2020bridging} & 39.2 & 57.4 & 42.2 \\
& \textbf{TOOD} & \textbf{42.5} & \textbf{59.8} & \textbf{46.4} \\
\cmidrule(r){1-1}
\cmidrule(r){2-2}
\cmidrule(r){3-5}
Anchor-based & ATSS~\cite{zhang2020bridging} & 39.3 & 57.5 & 42.8
\\
& \textbf{TOOD} & \textbf{42.4} & \textbf{59.8} & \textbf{46.1}
\\
\bottomrule
\end{tabular}
\caption{Performance of the complete TOOD~(\head + TAL).}
\label{ex-tuh-tsa}
\vspace{-2mm}
\end{table}

\begin{table}
\centering
\small
\begin{tabular}{llccc}
\toprule
$\alpha$ & $\beta$ & AP & AP$_{50}$ & AP$_{75}$ \\
\cmidrule(r){1-2}
\cmidrule(r){3-5}
0.5 & 2 & 42.4 & \textbf{60.0} & 46.1 \\
0.5 & 4 & 42.3 & 59.3 & 45.8 \\
0.5 & 6 & 41.7 & 58.1 & 45.1 \\
1.0 & 6  & \textbf{42.5} & 59.8 & \textbf{46.4} \\
1.0 & 8  & 42.2 & 59.0 & 46.0 \\
1.5 & 8 & 41.5 & 59.4 & 44.7
\\
\bottomrule
\end{tabular}
\caption{Analysis of different hyper-parameters for $t$.}
\label{ex-hyper-parameters-t}
\vspace{-4mm}
\end{table}

\begin{table*}[ht]
		\centering
		\small
		\begin{tabular}{lcccccccc}
			\toprule
			Method & Reference & Backbone & AP & AP$_{50}$ & AP$_{75}$ & AP$_{S}$ & AP$_{M}$ & AP$_{L}$ \\
			\cmidrule(r){1-1}
			\cmidrule(r){2-2}
			\cmidrule(r){3-3}
			\cmidrule(r){4-6}
			\cmidrule(r){7-9}
			RetinaNet~\cite{lin2017focal} & ICCV17 & ResNet-101 & 39.1 & 59.1 & 42.3 & 21.9 & 42.7 & 50.2 \\ 
			FoveaBox~\cite{kong2020foveabox} & - & ResNet-101 & 40.6 & 60.1 & 43.5 & 23.3 & 45.2 & 54.5 \\ 
			FCOS w/ imprv~\cite{tian2019fcos} & ICCV19 & ResNet-101 & 43.0 & 61.7 & 46.3 & 26.0 & 46.8 & 55.0 \\
			Noisy Anchor~\cite{li2020learning} & CVPR20 & ResNet-101 & 41.8 & 61.1 & 44.9 & 23.4 & 44.9 & 52.9 \\ 
			MAL~\cite{ke2020multiple} & CVPR20 & ResNet-101 & 43.6 & 62.8 & 47.1 & 25.0 & 46.9 & 55.8 \\
			SAPD~\cite{zhu2020soft} & CVPR20 & ResNet-101 & 43.5 & 63.6 & 46.5 & 24.9 & 46.8 & 54.6 \\
			ATSS~\cite{zhang2020bridging} & CVPR20 & ResNet-101 & 43.6 & 62.1 & 47.4 & 26.1 & 47.0 & 53.6 \\
			PAA~\cite{kim2020probabilistic} & ECCV20 & ResNet-101 & 44.8 & 63.3 & 48.7 & 26.5 & 48.8 & 56.3 \\
			GFL~\cite{li2020generalized} & NeurIPS20 & ResNet-101 & 45.0 & 63.7 & 48.9 & 27.2 & 48.8 & 54.5 \\
			\textbf{\name}~(ours) & - & ResNet-101 & \textbf{46.7} & \textbf{64.6} & \textbf{50.7} & \textbf{28.9} & \textbf{49.6} & \textbf{57.0} \\
			\cmidrule(r){1-1}
			\cmidrule(r){2-2}
			\cmidrule(r){3-3}
			\cmidrule(r){4-6}
			\cmidrule(r){7-9}
			SAPD~\cite{zhu2020soft} & CVPR20 & ResNeXt-101-64$\times$4d & 45.4 & 65.6 & 48.9 & 27.3 & 48.7 & 56.8 \\
			ATSS~\cite{zhang2020bridging} & CVPR20 & ResNeXt-101-64$\times$4d & 45.6 & 64.6 & 49.7 & 28.5 & 48.9 & 55.6 \\
			PAA~\cite{kim2020probabilistic} & ECCV20 & ResNeXt-101-64$\times$4d & 46.6 & 65.6 & 50.8 & 28.8 & 50.4 & 57.9 \\
			GFL~\cite{li2020generalized} & NeurIPS20 & ResNeXt-101-32$\times$4d & 46.0 & 65.1 & 50.1 & 28.2 & 49.6 & 56.0 \\
			\textbf{\name}~(ours) & - & ResNeXt-101-64$\times$4d & \textbf{48.3} & \textbf{66.5} & \textbf{52.4} & \textbf{30.7} & \textbf{51.3} & \textbf{58.6}
			\\
			\cmidrule(r){1-1}
			\cmidrule(r){2-2}
			\cmidrule(r){3-3}
			\cmidrule(r){4-6}
			\cmidrule(r){7-9}
			SAPD~\cite{zhu2020soft} & CVPR20 & ResNet-101-DCN & 46.0 & 65.9 & 49.6 & 26.3 & 49.2 & 59.6 \\
			ATSS~\cite{zhang2020bridging} & CVPR20 & ResNet-101-DCN & 46.3 & 64.7 & 50.4 & 27.7 & 49.8 & 58.4 \\
			PAA~\cite{kim2020probabilistic} & ECCV20 & ResNet-101-DCN & 47.4 & 65.7 & 51.6 & 27.9 & 51.3 & 60.6 \\
			GFL~\cite{li2020generalized} & NeurIPS20 & ResNet-101-DCN & 47.3 & 66.3 & 51.4 & 28.0 & 51.1 & 59.2 \\
			\textbf{\name}~(ours) & - & ResNet-101-DCN & \textbf{49.6} & \textbf{67.4} & \textbf{54.1} & \textbf{30.5} & \textbf{52.7} & \textbf{62.4}
			\\
			\cmidrule(r){1-1}
			\cmidrule(r){2-2}
			\cmidrule(r){3-3}
			\cmidrule(r){4-6}
			\cmidrule(r){7-9}
			SAPD~\cite{zhu2020soft} & CVPR20 & ResNeXt-101-64$\times$4d-DCN & 47.4 & 67.4 & 51.1 & 28.1 & 50.3 & 61.5 \\
			ATSS~\cite{zhang2020bridging} & CVPR20 & ResNeXt-101-64$\times$4d-DCN & 47.7 & 66.5 & 51.9 & 29.7 & 50.8 & 59.4 \\
			PAA~\cite{kim2020probabilistic} & ECCV20 & ResNeXt-101-64$\times$4d-DCN & 49.0 & 67.8 & 53.3 & 30.2 & 52.8 & 62.2 \\
			GFL~\cite{li2020generalized} & NeurIPS20 & ResNeXt-101-32$\times$4d-DCN & 48.2 & 67.4 & 52.6 & 29.2 & 51.7 & 60.2 \\
			GFLV2~\cite{li2021generalized}$^{\dag}$ & CVPR21 & ResNeXt-101-32$\times$4d-DCN & 49.0 & 67.6 & 53.5 & 29.7 & 52.4 & 61.4 \\
			OTA~\cite{ge2021ota}$^{\dag}$ & CVPR21 & ResNeXt-101-64$\times$4d-DCN & 49.2 & 67.6 & 53.5 & 30.0 & 52.5 & 62.3 \\
			IQDet~\cite{ma2021iqdet}$^{\dag}$ & CVPR21 & ResNeXt-101-64$\times$4d-DCN & 49.0 & 67.5 & 53.1 & 30.0 & 52.3 & 62.0 \\
			VFNet~\cite{zhang2021varifocalnet}$^{\dag}$ & CVPR21 & ResNeXt-101-64$\times$4d-DCN & 49.9 & 68.5 & 54.3 & 30.7 & 53.1 & 62.8 \\
			\textbf{\name}~(ours) & - & ResNeXt-101-64$\times$4d-DCN & \textbf{51.1} & \textbf{69.4} & \textbf{55.5} & \textbf{31.9} & \textbf{54.1} & \textbf{63.7}
			\\
			\bottomrule
		\end{tabular}
		\caption{Detection results on the COCO $test$-$dev$ set. $^{\dag}$ indicates the concurrent work.}
		\label{comparing}
		\vspace{-2mm}
\end{table*}

\subsection{Comparison with the State-of-the-Art}
\begin{table*}[ht]
		\centering
		\small
		\begin{tabular}{lcrrrrrrr}
			\toprule
			Method & AP & PCC~(top-50) & IoU~(top-10) & \#Correct boxes & \#Redundant boxes & \#Error boxes
			\\
			\cmidrule(r){1-1}
			\cmidrule(r){2-2}
			\cmidrule(r){3-4}
			\cmidrule(r){5-7}
			Parallel head + ATSS~\cite{zhang2020bridging} & 39.2 & 0.408 & 0.637 & 30,261 & 25,428 & 92,677
			\\
			\textbf{\head} + ATSS~\cite{zhang2020bridging} & 41.1 & 0.440 & 0.644 & 30,601 & 21,838 & 79,189
			\\
			Parallel head + \textbf{TAL} & 40.3 & 0.415 & 0.643 & 30,506 & 15,927 & 72,320
			\\
			\textbf{\head} + \textbf{TAL} & \textbf{42.5} & \textbf{0.452} & \textbf{0.661} & \textbf{30,734} & \textbf{15,242} & \textbf{69,013}
			\\
			\bottomrule
		\end{tabular}
		\caption{Analysis for task-alignment of TOOD with backbone ResNet-50.}
		\label{quantitative-analysis}
		\vspace{-3mm}
\end{table*}	

    We compare our TOOD with other one-stage detectors on the COCO $test$-$dev$ in Table~\ref{comparing}. The models are trained with scale jitter~(480-800) and for 2$\times$ learning schedule~(24 epochs) as the most current method~\cite{li2020generalized}. For a fair comparison, we report results of single model and single testing scale. With ResNet-101 and ResNeXt-101-64$\times$4d, \name achieves 46.7 AP and 48.3 AP, outperforming the most current one-stage detectors such as ATSS~\cite{zhang2020bridging}~(by $\sim$3 AP) and GFL~\cite{li2020generalized}~(by $\sim$2 AP).
    Furthermore, with ResNet-101-DCN and ResNeXt-101-64$\times$4d-DCN, \name brings a larger improvement, comparing to other detectors. For example, it obtains an improvement of 2.8 AP~(48.3$\rightarrow$51.1 AP) while ATSS has a 2.1 AP~(45.6$\rightarrow$47.7 AP) improvement. This validates that TOOD can cooperate with Deformable Convolutional Networks~(DCN)~\cite{zhu2019deformable} more efficiently, by adaptively adjusting the spatial distribution of the learned features for task-alignment. 
	Note that in TOOD, DCN is applied to the first two layers in the head tower.
	As shown in Table~\ref{comparing}, \name achieves a new state-of-the-art result with 51.1 AP in one-stage object detection.

\subsection{Quantitative Analysis for Task-alignment}
\vspace{-1mm}
	We quantitatively analyze the effect of the proposed methods on the alignment of two tasks.
	Without NMS, we calculate a Pearson Correlation Coefficient~(PCC) between the rankings~\cite{oksuz2020ranking} of classification and localization by selecting top-50 confident predictions for each instance, and a mean IoU of the top-10 confident predictions, averaged over instances. As shown in Table~\ref{quantitative-analysis}, the mean PCC and IoU are improved by using \head and TAL. 
	Meanwhile, with NMS, the number of the correct boxes~(IoU$>=$0.5) increases while those of the redundant~(IoU$>=$0.5) and error boxes~(0.1$<$IoU$<$0.5) decrease substantially when applying \head and TAL. The statistics suggest that TOOD is more compatible with NMS, by preserving more correct boxes, and suppressing the redundant/error boxes significantly. At last, detection performance is improved by 3.3 AP in total. Several detection examples are illustrated in Figure~\ref{illustration_of_images}.

\vspace{-2mm}
\section{Conclusion}
	In this work, we illustrate the misalignment between classification and localization in the existing one-stage detectors, and propose \name to align the two tasks. In particular, we design a task-aligned head to enhance the interaction of two tasks, and then improve its ability of learning the alignment. Furthermore, a new task-aligned learning strategy is developed by introducing a sample assignment scheme and new loss functions, both of which are computed via an anchor alignment metric. With these improvements, \name achieved a 51.1 AP on MS-COCO, surpassing the state-of-the-art one-stage detectors by a large margin.

{\small
\bibliographystyle{ieee_fullname}
\bibliography{egbib}

\begin{thebibliography}{10}\itemsep=-1pt

\bibitem{cao2020prime}
Yuhang Cao, Kai Chen, Chen~Change Loy, and Dahua Lin.
\newblock Prime sample attention in object detection.
\newblock In {\em Proceedings of the IEEE Conference on Computer Vision and
  Pattern Recognition}, pages 11583--11591, 2020.

\bibitem{deng2009imagenet}
Jia Deng, Wei Dong, Richard Socher, Li-Jia Li, Kai Li, and Li Fei-Fei.
\newblock Imagenet: A large-scale hierarchical image database.
\newblock In {\em Proceedings of the IEEE Conference on Computer Vision and
  Pattern Recognition}, pages 248--255, 2009.

\bibitem{duan2019centernet}
Kaiwen Duan, Song Bai, Lingxi Xie, Honggang Qi, Qingming Huang, and Qi Tian.
\newblock Centernet: Keypoint triplets for object detection.
\newblock In {\em Proceedings of the IEEE International Conference on Computer
  Vision}, pages 6569--6578, 2019.

\bibitem{feng2021exploring}
Chengjian Feng, Yujie Zhong, and Weilin Huang.
\newblock Exploring classification equilibrium in long-tailed object detection.
\newblock {\em arXiv preprint arXiv:2108.07507}, 2021.

\bibitem{ge2021ota}
Zheng Ge, Songtao Liu, Zeming Li, Osamu Yoshie, and Jian Sun.
\newblock Ota: Optimal transport assignment for object detection.
\newblock In {\em Proceedings of the IEEE Conference on Computer Vision and
  Pattern Recognition}, pages 303--312, 2021.

\bibitem{girshick2014rich}
Ross Girshick, Jeff Donahue, Trevor Darrell, and Jitendra Malik.
\newblock Rich feature hierarchies for accurate object detection and semantic
  segmentation.
\newblock In {\em Proceedings of the IEEE Conference on Computer Vision and
  Pattern Recognition}, pages 580--587, 2014.

\bibitem{jiang2018acquisition}
Borui Jiang, Ruixuan Luo, Jiayuan Mao, Tete Xiao, and Yuning Jiang.
\newblock Acquisition of localization confidence for accurate object detection.
\newblock In {\em Proceedings of the European Conference on Computer Vision},
  pages 784--799, 2018.

\bibitem{ke2020multiple}
Wei Ke, Tianliang Zhang, Zeyi Huang, Qixiang Ye, Jianzhuang Liu, and Dong
  Huang.
\newblock Multiple anchor learning for visual object detection.
\newblock In {\em Proceedings of the IEEE Conference on Computer Vision and
  Pattern Recognition}, pages 10206--10215, 2020.

\bibitem{kim2020probabilistic}
Kang Kim and Hee~Seok Lee.
\newblock Probabilistic anchor assignment with iou prediction for object
  detection.
\newblock In {\em Proceedings of the European Conference on Computer Vision},
  2020.

\bibitem{kong2020foveabox}
Tao Kong, Fuchun Sun, Huaping Liu, Yuning Jiang, Lei Li, and Jianbo Shi.
\newblock Foveabox: Beyound anchor-based object detection.
\newblock {\em IEEE Transactions on Image Processing}, 29:7389--7398, 2020.

\bibitem{law2018cornernet}
Hei Law and Jia Deng.
\newblock Cornernet: Detecting objects as paired keypoints.
\newblock In {\em Proceedings of the European Conference on Computer Vision},
  pages 734--750, 2018.

\bibitem{li2020learning}
Hengduo Li, Zuxuan Wu, Chen Zhu, Caiming Xiong, Richard Socher, and Larry~S
  Davis.
\newblock Learning from noisy anchors for one-stage object detection.
\newblock In {\em Proceedings of the IEEE Conference on Computer Vision and
  Pattern Recognition}, pages 10588--10597, 2020.

\bibitem{li2021generalized}
Xiang Li, Wenhai Wang, Xiaolin Hu, Jun Li, Jinhui Tang, and Jian Yang.
\newblock Generalized focal loss v2: Learning reliable localization quality
  estimation for dense object detection.
\newblock In {\em Proceedings of the IEEE Conference on Computer Vision and
  Pattern Recognition}, pages 11632--11641, 2021.

\bibitem{li2020generalized}
Xiang Li, Wenhai Wang, Lijun Wu, Shuo Chen, Xiaolin Hu, Jun Li, Jinhui Tang,
  and Jian Yang.
\newblock Generalized focal loss: Learning qualified and distributed bounding
  boxes for dense object detection.
\newblock In {\em Advances in Neural Information Processing Systems}, 2020.

\bibitem{lin2017feature}
Tsung-Yi Lin, Piotr Doll{\'a}r, Ross Girshick, Kaiming He, Bharath Hariharan,
  and Serge Belongie.
\newblock Feature pyramid networks for object detection.
\newblock In {\em Proceedings of the IEEE Conference on Computer Vision and
  Pattern Recognition}, pages 2117--2125, 2017.

\bibitem{lin2017focal}
Tsung-Yi Lin, Priya Goyal, Ross Girshick, Kaiming He, and Piotr Doll{\'a}r.
\newblock Focal loss for dense object detection.
\newblock In {\em Proceedings of the IEEE International Conference on Computer
  Vision}, pages 2980--2988, 2017.

\bibitem{lin2014microsoft}
Tsung-Yi Lin, Michael Maire, Serge Belongie, James Hays, Pietro Perona, Deva
  Ramanan, Piotr Doll{\'a}r, and C~Lawrence Zitnick.
\newblock Microsoft coco: Common objects in context.
\newblock In {\em Proceedings of the European Conference on Computer Vision},
  pages 740--755, 2014.

\bibitem{liu2016ssd}
Wei Liu, Dragomir Anguelov, Dumitru Erhan, Christian Szegedy, Scott Reed,
  Cheng-Yang Fu, and Alexander~C Berg.
\newblock Ssd: Single shot multibox detector.
\newblock In {\em Proceedings of the European Conference on Computer Vision},
  pages 21--37, 2016.

\bibitem{ma2021iqdet}
Yuchen Ma, Songtao Liu, Zeming Li, and Jian Sun.
\newblock Iqdet: Instance-wise quality distribution sampling for object
  detection.
\newblock In {\em Proceedings of the IEEE Conference on Computer Vision and
  Pattern Recognition}, pages 1717--1725, 2021.

\bibitem{oksuz2020ranking}
Kemal Oksuz, Baris~Can Cam, Emre Akbas, and Sinan Kalkan.
\newblock A ranking-based, balanced loss function unifying classification and
  localisation in object detection.
\newblock In {\em Advances in Neural Information Processing Systems}, 2020.

\bibitem{pang2019libra}
Jiangmiao Pang, Kai Chen, Jianping Shi, Huajun Feng, Wanli Ouyang, and Dahua
  Lin.
\newblock Libra r-cnn: Towards balanced learning for object detection.
\newblock In {\em Proceedings of the IEEE Conference on Computer Vision and
  Pattern Recognition}, pages 821--830, 2019.

\bibitem{redmon2016you}
Joseph Redmon, Santosh Divvala, Ross Girshick, and Ali Farhadi.
\newblock You only look once: Unified, real-time object detection.
\newblock In {\em Proceedings of the IEEE Conference on Computer Vision and
  Pattern Recognition}, pages 779--788, 2016.

\bibitem{ren2015faster}
Shaoqing Ren, Kaiming He, Ross Girshick, and Jian Sun.
\newblock Faster r-cnn: Towards real-time object detection with region proposal
  networks.
\newblock In {\em Advances in Neural Information Processing Systems}, pages
  91--99, 2015.

\bibitem{rezatofighi2019generalized}
Hamid Rezatofighi, Nathan Tsoi, JunYoung Gwak, Amir Sadeghian, Ian Reid, and
  Silvio Savarese.
\newblock Generalized intersection over union: A metric and a loss for bounding
  box regression.
\newblock In {\em Proceedings of the IEEE Conference on Computer Vision and
  Pattern Recognition}, pages 658--666, 2019.

\bibitem{sermanet2013overfeat}
Pierre Sermanet, David Eigen, Xiang Zhang, Micha{\"e}l Mathieu, Rob Fergus, and
  Yann LeCun.
\newblock Overfeat: Integrated recognition, localization and detection using
  convolutional networks.
\newblock {\em arXiv preprint arXiv:1312.6229}, 2013.

\bibitem{song2020revisiting}
Guanglu Song, Yu Liu, and Xiaogang Wang.
\newblock Revisiting the sibling head in object detector.
\newblock In {\em Proceedings of the IEEE Conference on Computer Vision and
  Pattern Recognition}, pages 11563--11572, 2020.

\bibitem{tian2019fcos}
Zhi Tian, Chunhua Shen, Hao Chen, and Tong He.
\newblock Fcos: Fully convolutional one-stage object detection.
\newblock In {\em Proceedings of the IEEE International Conference on Computer
  Vision}, pages 9627--9636, 2019.

\bibitem{wu2019double}
Yue Wu, Yinpeng Chen, Lu Yuan, Zicheng Liu, Lijuan Wang, Hongzhi Li, and Yun
  Fu.
\newblock Double-head rcnn: Rethinking classification and localization for
  object detection.
\newblock {\em arXiv preprint arXiv:1904.06493}, 2, 2019.

\bibitem{zhang2020localize}
Heng Zhang, Elisa Fromont, S{\'e}bastien Lef{\`e}vre, and Bruno Avignon.
\newblock Localize to classify and classify to localize: Mutual guidance in
  object detection.
\newblock In {\em Proceedings of the Asian Conference on Computer Vision},
  2020.

\bibitem{zhang2021varifocalnet}
Haoyang Zhang, Ying Wang, Feras Dayoub, and Niko Sunderhauf.
\newblock Varifocalnet: An iou-aware dense object detector.
\newblock In {\em Proceedings of the IEEE Conference on Computer Vision and
  Pattern Recognition}, pages 8514--8523, 2021.

\bibitem{zhang2020bridging}
Shifeng Zhang, Cheng Chi, Yongqiang Yao, Zhen Lei, and Stan~Z Li.
\newblock Bridging the gap between anchor-based and anchor-free detection via
  adaptive training sample selection.
\newblock In {\em Proceedings of the IEEE Conference on Computer Vision and
  Pattern Recognition}, pages 9759--9768, 2020.

\bibitem{zhang2019freeanchor}
Xiaosong Zhang, Fang Wan, Chang Liu, Rongrong Ji, and Qixiang Ye.
\newblock Freeanchor: Learning to match anchors for visual object detection.
\newblock In {\em Advances in Neural Information Processing Systems}, pages
  147--155, 2019.

\bibitem{zhong2020representation}
Yujie Zhong, Zelu Deng, Sheng Guo, Matthew~R Scott, and Weilin Huang.
\newblock Representation sharing for fast object detector search and beyond.
\newblock In {\em Proceedings of the European Conference on Computer Vision},
  pages 471--487, 2020.

\bibitem{zhou2019objects}
Xingyi Zhou, Dequan Wang, and Philipp Kr{\"a}henb{\"u}hl.
\newblock Objects as points.
\newblock {\em arXiv preprint arXiv:1904.07850}, 2019.

\bibitem{zhu2020soft}
Chenchen Zhu, Fangyi Chen, Zhiqiang Shen, and Marios Savvides.
\newblock Soft anchor-point object detection.
\newblock In {\em Proceedings of the European Conference on Computer Vision},
  2020.

\bibitem{zhu2019feature}
Chenchen Zhu, Yihui He, and Marios Savvides.
\newblock Feature selective anchor-free module for single-shot object
  detection.
\newblock In {\em Proceedings of the IEEE Conference on Computer Vision and
  Pattern Recognition}, pages 840--849, 2019.

\bibitem{zhu2019deformable}
Xizhou Zhu, Han Hu, Stephen Lin, and Jifeng Dai.
\newblock Deformable convnets v2: More deformable, better results.
\newblock In {\em Proceedings of the IEEE Conference on Computer Vision and
  Pattern Recognition}, pages 9308--9316, 2019.

\end{thebibliography}
}

\clearpage
\includepdf[pages=1]{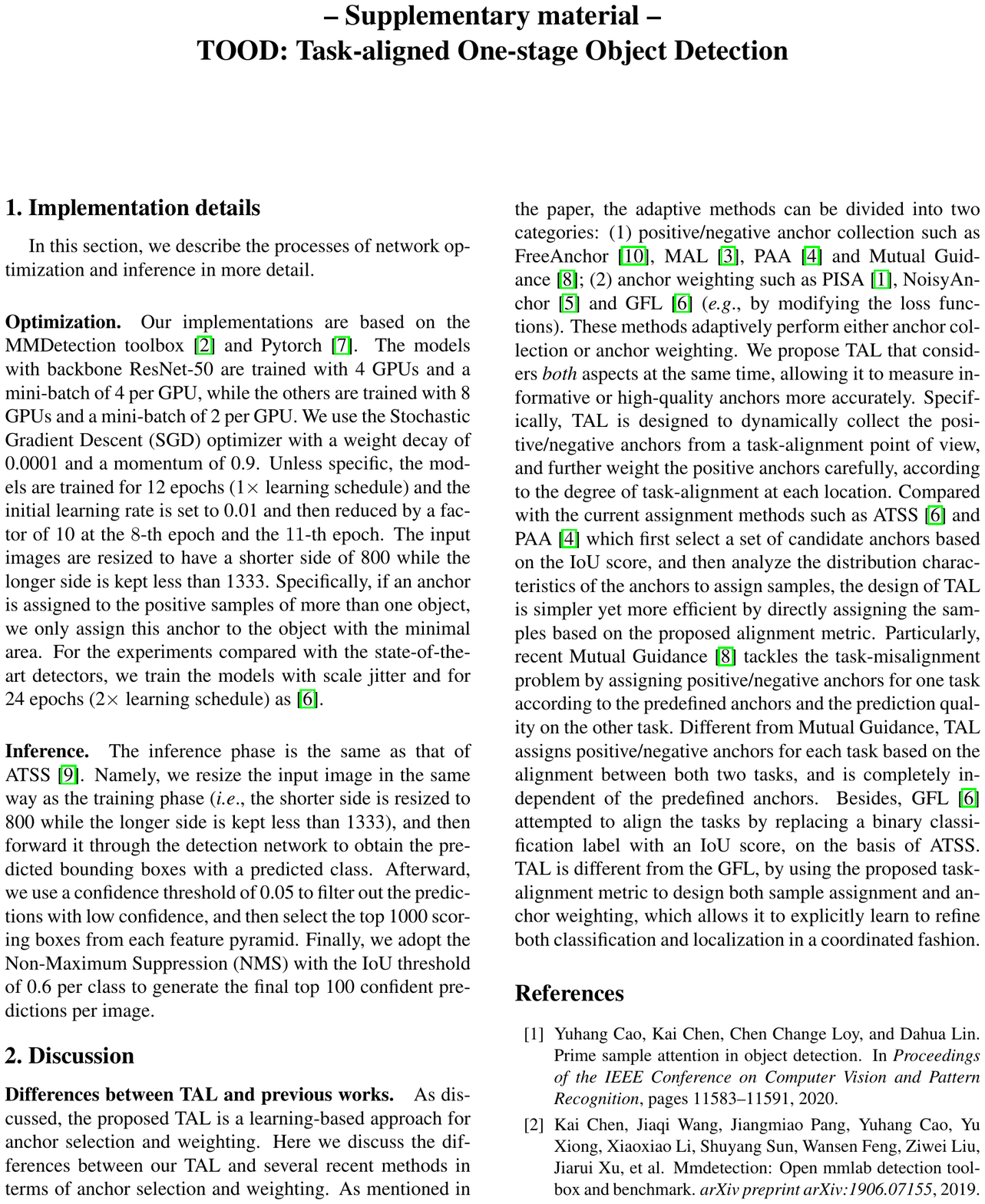}
\includepdf[pages=2]{sup.pdf}

\end{document}